\documentclass[conference]{IEEEtran}
\IEEEoverridecommandlockouts

\usepackage{cite}
\usepackage{amsmath,amssymb,amsfonts}
\usepackage{algorithmic}
\usepackage{graphicx}
\usepackage{textcomp}
\usepackage{xcolor}
\def\BibTeX{{\rm B\kern-.05em{\sc i\kern-.025em b}\kern-.08em
    T\kern-.1667em\lower.7ex\hbox{E}\kern-.125emX}}
\begin{document}

\makeatletter
\newcommand{\linebreakand}{%
  \end{@IEEEauthorhalign}
  \hfill\mbox{}\par
  \mbox{}\hfill\begin{@IEEEauthorhalign}
}
\makeatother

\title{BatteryBERT for Realistic Battery Fault Detection Using Point-Masked Signal Modeling\\
\thanks{This work is supported by the National Natural Science Foundation of China (62273197 and 62403276), and the Beijing Natural Science Foundation under Grant L233027.}
}

\author{\IEEEauthorblockN{Songqi Zhou}
\IEEEauthorblockA{\textit{Department of Automation} \\
\textit{Tsinghua University}\\
Beijing, China \\
zhousongqi@mail.tsinghua.edu.cn}
\and
\IEEEauthorblockN{Ruixue Liu}
\IEEEauthorblockA{\textit{Department of Automation} \\
\textit{Tsinghua University}\\
Beijing, China \\
liuruixue@mail.tsinghua.edu.cn}
\and
\IEEEauthorblockN{Yixing Wang}
\IEEEauthorblockA{\textit{Department of Automation} \\
\textit{Tsinghua University}\\
Beijing, China \\
yx-wang21@mails.tsinghua.edu.cn}
\linebreakand 
\IEEEauthorblockN{Jia Lu}
\IEEEauthorblockA{\textit{Department of Automation} \\
\textit{Tsinghua University}\\
Beijing, China \\
lu-j22@mails.tsinghua.edu.cn}
\and
\IEEEauthorblockN{Benben Jiang*}
\IEEEauthorblockA{\textit{Department of Automation} \\
\textit{Tsinghua University}\\
Beijing, China \\
bbjiang@tsinghua.edu.cn}
}

\maketitle

\begin{abstract}
Accurate fault detection in lithium-ion batteries is essential for the safe and reliable operation of electric vehicles and energy storage systems. However, existing methods often struggle to capture complex temporal dependencies and cannot fully leverage abundant unlabeled data. Although large language models (LLMs) exhibit strong representation capabilities, their architectures are not directly suited to the numerical time-series data common in industrial settings. To address these challenges, we propose a novel framework that adapts BERT-style pretraining for battery fault detection by extending the standard BERT architecture with a customized time-series-to-token representation module and a point-level Masked Signal Modeling (point-MSM) pretraining task tailored to battery applications. This approach enables self-supervised learning on sequential current, voltage, and other charge–discharge cycle data, yielding distributionally robust, context-aware temporal embeddings. We then concatenate these embeddings with battery metadata and feed them into a downstream classifier for accurate fault classification. Experimental results on a large-scale real-world dataset show that models initialized with our pretrained parameters significantly improve both representation quality and classification accuracy, achieving an AUROC of 0.945 and substantially outperforming existing approaches. These findings validate the effectiveness of BERT-style pretraining for time-series fault detection.
\end{abstract}


\begin{figure*}[ht]
    \centering
    \includegraphics[width=0.9\textwidth]{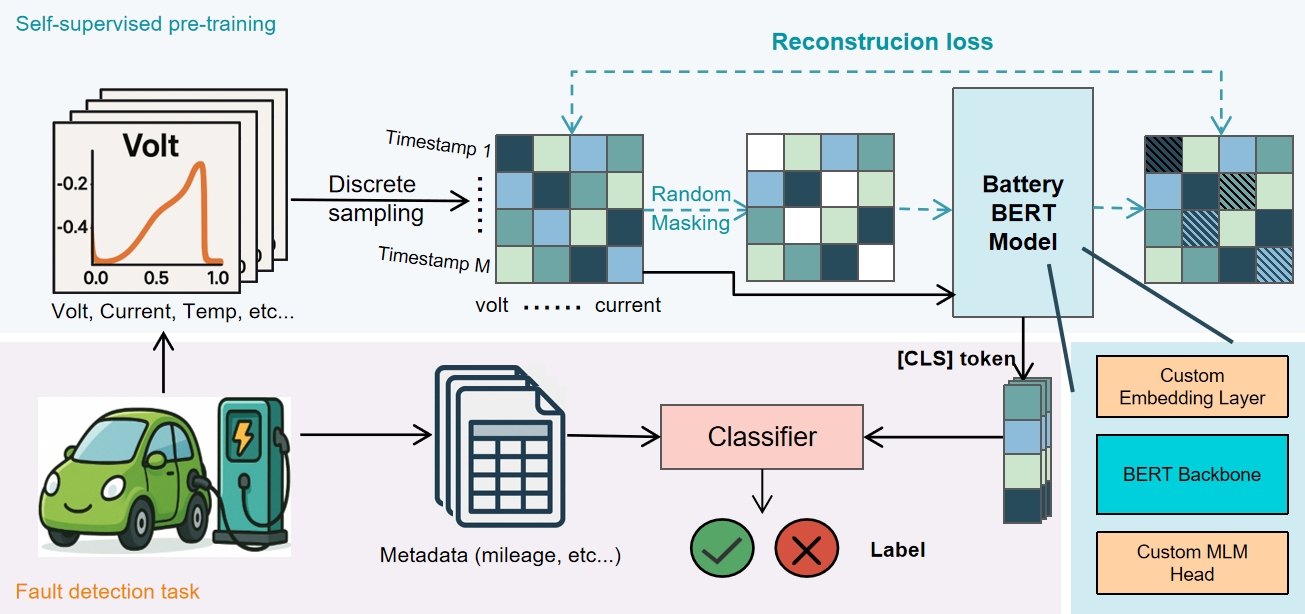}
    \caption{The \textbf{BatteryBERT} framework architecture. The model is first pre-trained on multivariate time-series signals using a point-MSM pretext task. For downstream fault detection, the model extracts a dynamic sequence embedding (\texttt{[CLS]} token), which is then fused with static metadata and input to a classifier for final fault prediction.}
    \label{fig:1}
\end{figure*}

\section{Introduction}
Ensuring the reliable and safe operation of lithium-ion batteries is critical for the continued development of electric vehicles (EVs). However, the increasing integration of batteries into safety-critical EV systems places growing demands on timely and accurate fault detection. Undetected faults can escalate into catastrophic failures, such as sudden capacity loss or thermal runaway, posing serious risks to both vehicle performance and user safety.

Recent advances in machine learning have improved battery fault diagnosis. For example, Zhang et al.\cite{ref6} employed a dynamic autoencoder to model input–response mappings and detect early faults through reconstruction errors. Sun et al.\cite{ref7} proposed an autoencoder-enhanced regularized prototypical network, combining pretrained autoencoders with a regularized embedding strategy to address class imbalance. Lai et al.\cite{ref8} developed an early warning approach for thermal runaway based on internal pressure trends in lithium iron phosphate (LFP) batteries. Cao et al.\cite{ref9} introduced a model-constrained deep learning framework that allows for real-time, multi-fault diagnosis under stochastic conditions. He et al.\cite{ref10} presented a model-data dual-driven method using long-short-term-memory networks within a cyber-physical framework for real-time fault prediction. Zhang et al.\cite{ref11} proposed a graph-guided ensemble learning approach to improve fault detection accuracy and recall in multi-type EV batteries. Conventional methods may struggle to capture the complex temporal dynamics of battery operations in electric vehicles.

In addition, the rarity of fault events limits the availability of labeled data, reducing the effectiveness of supervised learning. Meanwhile, large volumes of unlabeled time-series data - recorded during routine charge-discharge cycles - remain underutilized\cite{ref5}, representing a missed opportunity to improve detection accuracy.

Large language models (LLMs) have revolutionized natural language processing (NLP) through self-supervised pretraining, enabling the extraction of context-rich representations from massive unlabeled corpora\cite{min2023recent}. Inspired by this success, researchers have begun to adapt LLMs to industrial time-series tasks. However, these models are not natively designed for numerical multivariate signals, and their application to battery fault detection remains underexplored.

In this field, several studies have explored the use of LLMs for battery diagnosis. Zhao et al.\cite{ref1} proposed BatteryGPT, which uses retrieval augmented generation (RAG) for expert-level knowledge synthesis on fast-charging technologies. Peng et al.\cite{ref2} proposed an Internet of Batteries (IoB) framework enhanced by LLMs for electric vehicle battery health management. In \cite{ref3} and \cite{ref4}, Bian et al. proposed a hybrid prompt-driven LLM approach for robust SOC estimation of multi-type Li-ion batteries, achieving high accuracy and robustness under diverse operating conditions. Zhang et al.\cite{ref5} proposed a multi-cycle charging information-guided SOH estimation method for lithium-ion batteries, leveraging a pre-trained LLM and feature combinations to achieve high accuracy and stability in battery state estimation.

Building on these foundations, we propose a novel LLM-based framework for EV battery fault diagnosis. Our method extends the BERT\cite{ref_bert_variant} architecture with a customized time-series-to-token representation module to process multivariate sensor data. Via a point-level Masked Signal Modeling (point-MSM) pretext task, the model is pretrained on unlabeled current and voltage sequences to learn context-aware representations of battery behavior. We then combine these embeddings with metadata and feed them into a classifier for precise fault detection. Experiments on a large-scale real-world EV dataset demonstrate that our approach significantly outperforms existing methods, achieving an AUROC of 0.945. This highlights the unique advantages of BERT-style pretraining for battery time-series data: achieving strong generalization under limited labels, 
providing distributionally robust temporal modeling, and enabling seamless integration of domain knowledge with raw sensor data.

In summary, our work offers an efficient, scalable solution for EV battery fault detection and, more broadly, demonstrates the potential of adapting LLMs to structured industrial time-series tasks for predictive maintenance.

\begin{figure}[ht]
    \centering
    \includegraphics[width=0.9\columnwidth]{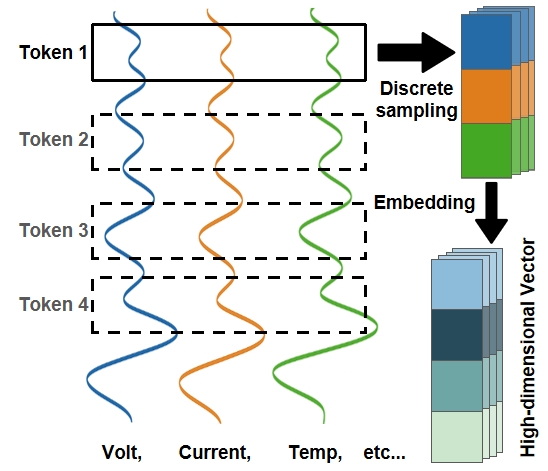}
    \caption{Tokenization and embedding pipeline for time-series data. Multivariate signals are discretized into a sequence of feature vectors (tokens). Each token is then projected into a high-dimensional embedding space for model input.}
    \label{fig:3}
\end{figure}

\section{Proposed Methodology}

To address the limitations of existing methods in capturing the complex long-range temporal dependencies inherent in battery operational data, we introduce \textbf{BatteryBERT}, a novel framework for realistic battery fault detection. This framework adapts the powerful BERT-style pretraining paradigm to the unique characteristics of battery time-series. Our core hypothesis is that a large-scale model can learn distributionally robust, context-aware temporal representations through self-supervised learning on vast unlabeled datasets. These deep features are then synergistically combined with static battery metadata for precise fault classification. As depicted in Figure~\ref{fig:1}, the overall framework comprises three key stages:(i) time-series representation and model architecture design tailored for numerical battery signals, (ii) self-supervised pre-training based on the point-MSM strategy to capture context-aware temporal dependencies, and (iii) supervised fine-tuning for the downstream fault detection task.

\subsection{Model Architecture and Time-Series Representation}

To effectively bridge the modality gap between continuous numerical signals and the BERT model architecture, we designed a specific data representation pipeline and tailored network modules, as illustrated in Figure~\ref{fig:3}.

\subsubsection{Signal Preprocessing and Tokenization}

As a preliminary step, all $D$ physical measurement channels are subjected to \textit{z}-score normalization. An entire operational cycle is represented as a sequence of $M$ feature vectors, $S = (\mathbf{x}_1, \mathbf{x}_2, \dots, \mathbf{x}_M)$. In our framework, each time-stamped measurement vector $\mathbf{x}_t = [V_t, I_t, T_t, \dots]^\top \in \mathbb{R}^D$ is treated as a single input token. The model receives a tensor of shape \texttt{[batch\_size, sequence\_length, num\_features]}.

\subsubsection{Time-Series Embedding Module}
\label{sec:embedding}
The embedding module transforms the raw numerical input sequence into a sequence of high-dimensional vectors suitable for the Transformer backbone. This process involves three sequential steps:
\begin{enumerate}
    \item \textbf{Feature Projection:} Each input token $\mathbf{x}_t \in \mathbb{R}^D$ is projected into the model's hidden dimension $H$ via a learnable linear layer.
    \begin{equation}
        \mathbf{h}_t^{(\text{proj})} = \mathbf{x}_t \mathbf{W}_e + \mathbf{b}_e
    \end{equation}
    where $\mathbf{W}_e \in \mathbb{R}^{D \times H}$ and $\mathbf{b}_e \in \mathbb{R}^H$ are the weights and bias of the projection layer, respectively.

    \item \textbf{Positional Encoding:} To inject temporal order, a learnable absolute positional embedding $\mathbf{p}_t \in \mathbb{R}^H$ is added to the projected vector.
    \begin{equation}
        \mathbf{h}_t^{(\text{pos})} = \mathbf{h}_t^{(\text{proj})} + \mathbf{p}_t
    \end{equation}

    \item \textbf{Normalization and Dropout:} Following standard Transformer practice, the resulting sum is processed by a Layer Normalization step, followed by a Dropout layer to yield the final embedding $\mathbf{e}_t$.
    \begin{equation}
        \mathbf{e}_t = \text{Dropout}(\text{LayerNorm}(\mathbf{h}_t^{(\text{pos})}))
    \end{equation}
\end{enumerate}
The resulting sequence of embeddings $\mathbf{E} = (\mathbf{e}_1, \dots, \mathbf{e}_M)$ serves as the input to the main encoder.

\subsubsection{BERT Backbone}
The core of our framework is a powerful Transformer encoder built upon the \textbf{BERT-base} architecture. Specifically, it comprises 12 stacked Transformer layers, each featuring a hidden dimension of \(H = 768\) and 12 self-attention heads. With approximately 110 million parameters, this deep and wide model has the capacity to capture subtle relationships in complex battery data across diverse operating conditions and vehicle types.

\subsubsection{Reconstruction Head for Pre-training}
During pre-training, the model's objective is to reconstruct the original numerical signals. To this end, we designed a simple linear regression head, denoted as $g_{\text{head}}$. This head takes the final hidden state $\mathbf{h}'_t \in \mathbb{R}^H$ from the BERT backbone and maps it back to the original feature dimension $D$. The reconstructed feature vector $\hat{\mathbf{x}}_t$ is given by:
\begin{equation}
    \hat{\mathbf{x}}_t = g_{\text{head}}(\mathbf{h}'_t) = \mathbf{h}'_t \mathbf{W}_h + \mathbf{b}_h
    \label{eq:reconstruction_head}
\end{equation}
where $\mathbf{W}_h \in \mathbb{R}^{H \times D}$ and $\mathbf{b}_h \in \mathbb{R}^D$ are the learnable parameters of the head. The output is a sequence of reconstructed vectors $\hat{S} = (\hat{\mathbf{x}}_1, \dots, \hat{\mathbf{x}}_M)$.

\subsection{Self-Supervised Pre-training: point-MSM}

We adopt a \textbf{point-MSM} task for self-supervised pre-training. Unlike traditional token-level masking, this strategy randomly omits individual feature dimensions within each time-series token and then regresses to recover their true values. Here, a token represents a multidimensional observation—such as voltage, current, and temperature—and at each time step we introduce $F$ feature-wise mask positions. By masking individual dimensions rather than entire tokens, the model is explicitly encouraged to capture complex inter-dimensional relationships. Moreover, because each time step yields $F$ separate reconstruction targets, the total number of training signals increases by a factor of $F$. This fine-grained masking scheme substantially boosts both training efficiency and the quality of learned representations, making it particularly well suited for pre-training large-scale models.

We define a binary mask matrix $\mathcal{M} \in \{0, 1\}^{M \times D}$, where an element $\mathcal{M}_{t,d}=1$ indicates that the $d$-th feature of the token at time $t$ is to be masked, and $\mathcal{M}_{t,d}=0$ otherwise. Approximately 15\% of the total features in a sequence are randomly selected for masking. The input sequence $S$ is then corrupted by replacing the masked values with zero, creating $S_{\text{masked}}$. This point-wise masking scheme compels the model to learn both inter-timestamp (temporal) dynamics and intra-timestamp (cross-feature) correlations.

The model takes $S_{\text{masked}}$ as input and produces a reconstructed sequence $\hat{S}$. The pre-training objective is to minimize the reconstruction loss $\mathcal{L}_{\text{MSM}}$, defined as the Mean Squared Error (MSE) calculated only over the masked elements. This is formally expressed using the Hadamard (element-wise) product $\odot$ and the Frobenius norm:
\begin{equation}
    \mathcal{L}_{\text{MSM}} = \frac{\| \mathcal{M} \odot (\hat{S} - S) \|_F^2}{\| \mathcal{M} \|_1}
    \label{eq:msm_loss}
\end{equation}
where $\| \mathcal{M} \|_1$ is the L1 norm, which simply counts the total number of masked elements, ensuring the loss is a properly scaled mean.

\subsection{Supervised Fine-tuning for Fault Detection}

Once pre-trained, the \textbf{BatteryBERT} model is fine-tuned for the downstream fault detection task. This stage integrates the learned dynamic features with static information for robust classification.

\subsubsection{Dynamic and Static Feature Engineering}
A full, unmasked time-series sequence $S$, prepended with a special \texttt{[CLS]} token, is fed into the pre-trained model. The final hidden state corresponding to the \texttt{[CLS]} token, $\mathbf{h}'_{\texttt{[CLS]}} \in \mathbb{R}^H$, is extracted to serve as a holistic, aggregated representation of the sequence's dynamics. We utilize this as the dynamic feature vector, $\mathbf{v}_{\text{temporal}}$:
\begin{equation}
    \mathbf{v}_{\text{temporal}} = \mathbf{h}'_{\texttt{[CLS]}}
\end{equation}
To create a comprehensive feature set, we concatenate this dynamic feature vector with a static metadata vector $\mathbf{v}_{\text{meta}} \in \mathbb{R}^K$, which includes attributes such as cumulative mileage and cycle count. The final fused feature vector $\mathbf{v}_{\text{final}}$ is formed as:
\begin{equation}
    \mathbf{v}_{\text{final}} = [\mathbf{v}_{\text{temporal}} ; \mathbf{v}_{\text{meta}}]
\end{equation}
where $[;]$ denotes the concatenation operation, resulting in a vector $\mathbf{v}_{\text{final}} \in \mathbb{R}^{H+K}$.

\subsubsection{Classification Head}
Finally, the fused feature vector $\mathbf{v}_{\mathrm{final}}$ is passed to a downstream classifier $C$. While end-to-end fine-tuning with a neural network head often yields superior accuracy, our objective in this study is to evaluate the representational quality of the extracted features. Accordingly, we adopt \textbf{LightGBM}\cite{ke2017lightgbm}, a high-efficiency gradient-boosting framework, as our classifier. This choice permits an unconfounded assessment of feature efficacy without the additional complexity of tuning a deep-learning head. Given $\mathbf{v}_{\mathrm{final}}$, the classifier predicts a fault probability
$\hat{y} = C(\mathbf{v}_{\mathrm{final}})$,
which is compared against the ground-truth label $y \in \{0,1\}$. Model parameters are learned by minimizing a standard binary cross-entropy loss.

\section{Experimental Setup}
This section details the dataset, experimental design, and evaluation metrics used to validate our proposed model and compare its performance against existing methods.
\subsection{Dataset}
For our experiments, we utilize the large-scale, real-world electric vehicle battery dataset introduced by Zhang et al \cite{ref6}. This public dataset was curated to facilitate the development and validation of fault detection algorithms in realistic settings. The dataset is comprehensive, comprising over \textbf{690,000 charging snippets collected from 347 distinct EVs}. It includes 292 vehicles labeled as ``normal'' and 55 labeled as ``abnormal'' (i.e., having a confirmed Li-ion battery fault). Each charging snippet contains critical time-series data from the Battery Management System (BMS), such as current, voltage, and temperature. By using this established benchmark dataset, we ensure our results are directly comparable to those reported in recent state-of-the-art studies.

\subsection{Experimental Design}

To ensure a fair and robust comparison, our experimental design adheres to the validation protocol established by prior studies on this dataset. Data are partitioned at the vehicle level into training and validation sets, such that all charging snippets from any given vehicle are assigned exclusively to either the training set or the validation set. This vehicle‐level split prevents data leakage and guarantees that model evaluation measures generalization to unseen vehicles rather than merely unseen charging cycles from familiar vehicles. Furthermore, during pre-training, only the training set was utilized, while the validation set was entirely withheld from all training procedures to ensure unbiased model assessment and to prevent any inadvertent leakage of information.

\subsection{Evaluation Metrics}

We assess the performance of our framework using a combination of quantitative and qualitative metrics to provide a comprehensive evaluation. These metrics are consistent with those used in the reference paper.

\begin{itemize}
    \item \textbf{AUROC:} 
    Our primary metric is the AUROC, which evaluates the model’s ability to discriminate between normal and abnormal classes across all thresholds—higher values denote stronger predictive performance.

    \item \textbf{t-SNE Visualization:} To qualitatively evaluate the quality of the learned representations, we use t-distributed Stochastic Neighbor Embedding (t-SNE). This technique projects the high-dimensional embeddings of charging snippets onto a 2D plane, allowing for visual inspection of how well the model separates different classes and normalizes data from different vehicles\cite{cheng2021supervised}.

    \item \textbf{Average Direct Cost:} To measure the practical and economic viability of our model, we calculate the expected direct cost, following the methodology and formula proposed by Zhang et al.\cite{ref6}. This cost function balances the direct costs of a battery fault ($c_f$) against the costs of inspection ($c_r$), weighted by the model's true positive ($q_{TP}$) and false positive ($q_{FP}$) rates, as well as the statistical fault rate ($p$). The formula is expressed as:
    \begin{equation}
    \begin{split}
        y(p, c_f, c_r, q_{TP}, q_{FP}) = & p(1 - q_{TP})c_f \\
        & + [pq_{TP} + (1 - p)q_{FP}]c_r
    \end{split}
    \end{equation}
    For this calculation, we adopt the following empirical cost and rate statistics from the reference paper:
    \begin{itemize}
        \item The EV battery fault rate ($p$) is \textbf{0.038\%}.
        \item The direct cost of a battery fault ($c_f$) is \textbf{5 million CNY} per vehicle.
        \item The direct cost of an inspection ($c_r$) is \textbf{8 thousand CNY} per vehicle.
    \end{itemize}
    This metric provides a crucial assessment of a model's real-world economic impact and effectiveness.
\end{itemize}

\section{Results and Analysis}

To validate the effectiveness of our proposed \textbf{BatteryBERT} framework, we conducted a series of three comprehensive experiments. These experiments were designed to: (i) evaluate the benefits of leveraging pre-trained parameters from NLP tasks for initializing our model; (ii) visualize and assess the quality of the temporal representations learned by \textbf{BatteryBERT} through self-supervised pre-training; and (iii) benchmark the performance of our method against existing state-of-the-art approaches on a downstream fault detection task.

\subsection{Experiment 1: Efficacy of NLP Pre-trained Initialization}

In this experiment, we investigated the impact of model initialization on the convergence and performance of the point-MSM pre-training task. We compared two scenarios: one where the model's weights were initialized using parameters from a standard BERT model pre-trained on a large text corpus, and another where the weights were initialized randomly. Both models were then pre-trained on our battery time-series dataset using the point-MSM objective.

\begin{figure}[ht]
    \centering
    \includegraphics[width=\columnwidth]{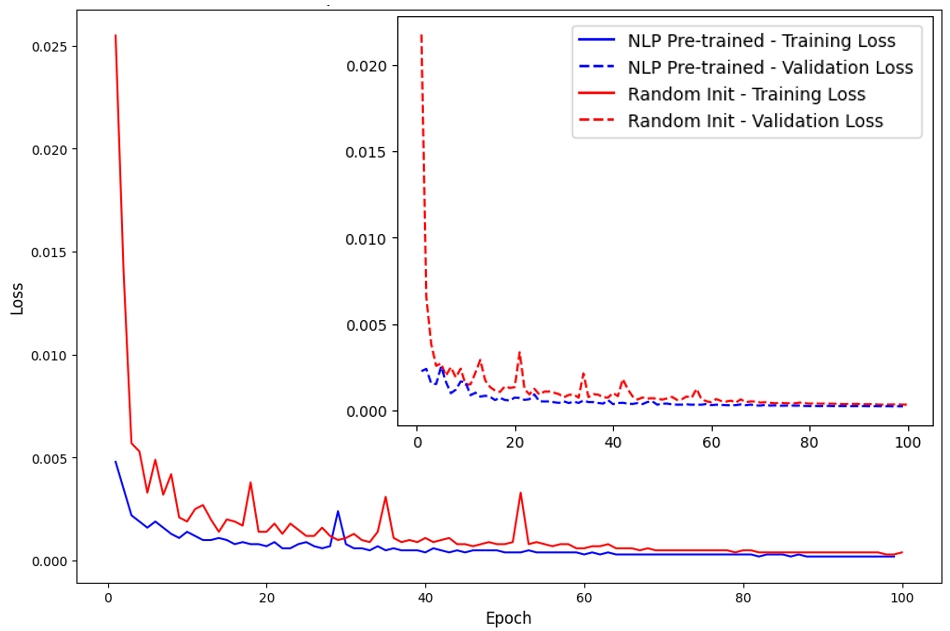}
    \caption{The effect of initialization on pre-training convergence. Compared to random initialization (red), using NLP pre-trained weights (blue) results in a significantly lower initial loss, faster convergence, and a more stable training process on the point-MSM pretext task.}
    \label{fig:2}
\end{figure}

As shown in Figure~\ref{fig:2}, the benefits of using NLP pre-trained parameters are evident. The \textbf{NLP Pre-trained model (blue lines)} starts with a significantly lower training and validation loss compared to the \textbf{Randomly Initialized model (red lines)}. Furthermore, it converges much faster and to a lower final loss value, exhibiting greater stability throughout the training process. In contrast, the randomly initialized model suffers from a high initial loss and demonstrates considerable instability, as indicated by the frequent spikes in its loss curves. This result confirms our hypothesis that knowledge transfer from the language domain provides a superior starting point for learning representations of numerical time-series, accelerating convergence and leading to a better final model for the point-MSM pretext task.

\subsection{Experiment 2: Representation Quality of Pre-trained Embeddings}

To qualitatively assess the representations learned by our pre-trained \textbf{BatteryBERT} model, we visualized the embeddings of time-series data from two distinct electric vehicles, denoted as EV1 and EV2. We used the t-SNE algorithm to project the high-dimensional representations into a two-dimensional space.

\begin{figure}[ht]
    \centering
    \includegraphics[width=\columnwidth]{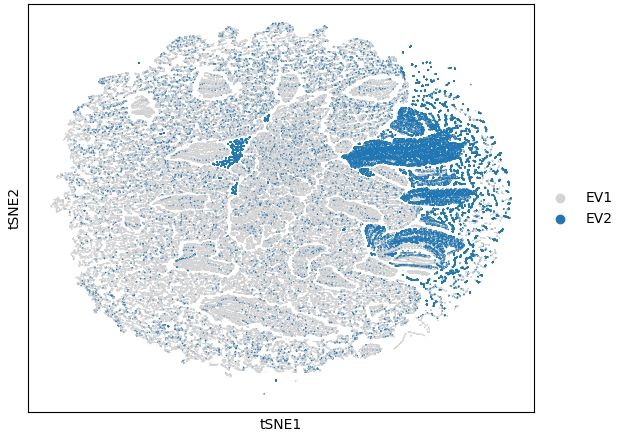}
    \caption{t-SNE visualization of raw data features before pre-training.}
    \label{fig:4}
\end{figure}

\begin{figure}[ht]
    \centering
    \includegraphics[width=\columnwidth]{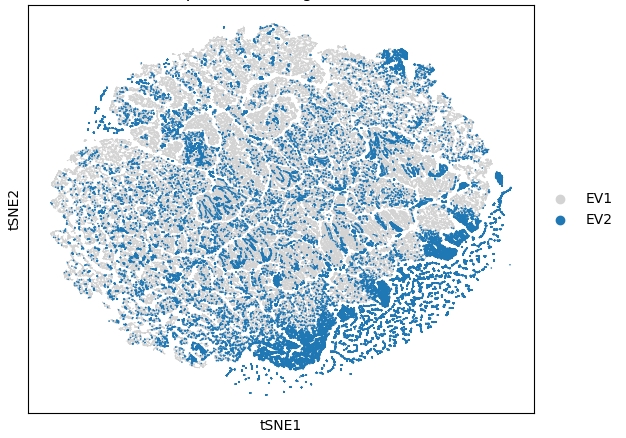}
    \caption{t-SNE visualization of BatteryBERT embeddings after pre-training.}
    \label{fig:5}
\end{figure}

The results, presented in Figure~\ref{fig:4} and~\ref{fig:5}, reveal a remarkable improvement in distributional alignment after pre-training. The first plot shows the distribution of the \textbf{raw, unprocessed data}. Here, the data points from EV1 (light grey) and EV2 (blue) form largely separate and distinct clusters, indicating a significant distributional difference between the two vehicles. This variance could stem from differences in manufacturing, usage patterns, or aging, posing a challenge for a generalizable fault detection model.

The second plot shows the distribution of the same data after being encoded by the pre-trained \textbf{BatteryBERT}. The representations for EV1 and EV2 are now substantially more intermingled, approaching a \textbf{more uniform co-distribution}. This demonstrates that the model has successfully learned to capture the fundamental underlying dynamics of battery operation while abstracting away vehicle-specific idiosyncrasies. By mapping heterogeneous data to a common representational space, \textbf{BatteryBERT} creates more distributionally robust and generalizable features, which is crucial for building high-performance downstream models.

\subsection{Experiment 3: Downstream Fault Detection Performance}


The ultimate goal of our framework is to improve the accuracy of battery fault detection. We evaluated the performance of \textbf{BatteryBERT} on a real-world fault detection dataset and compared it gainst several established baseline methods. These include Dynamical Deep Learning (DyAD) \cite{ref6}, Graph Deviation Network (GDN)\cite{deng2021graph}, Autoencoder (AE), Support Vector Data Description (SVDD), Gaussian Process (GP), and Variation Evaluation (VE)\cite{stent2005evaluating}.

\begin{table}[htbp]
\caption{Algorithm Performance and Cost Analysis}
\begin{center}
\begin{tabular}{|c|c|c|}
\hline
\textbf{Algorithm} & \textbf{AUROC (\%)}& \textbf{Average Direct Cost (CNY)} \\
\hline
BatteryBERT & \textbf{94.5} & \textbf{229} \\
\hline
DyAD & 88.6 & 850 \\
\hline
GDN & 70.3 & 1260 \\
\hline
AE & 72.8 & 1330 \\
\hline
SVDD & 51.5 & 1520 \\
\hline
GP & 66.6 & 1620 \\
\hline
VE & 55.6 & 1690 \\
\hline
\end{tabular}
\label{tab:performance_cost}
\end{center}
\end{table}

The results, summarized in Table~\ref{tab:performance_cost}, unequivocally demonstrate the superiority of our proposed method. \textbf{BatteryBERT achieves an AUROC of 94.5\%}, significantly outperforming the next best method, DyAD (88.6\%), by a substantial margin. All other baseline methods fall considerably short, with AUROC scores ranging from 51.5\% to 72.8\%.

Moreover, the table highlights a critical practical advantage of our approach. The ``Average Direct Cost'' associated with a fault, which can be interpreted as a metric of the model's precision and real-world financial impact, is lowest for \textbf{BatteryBERT} at just 229 CNY. This is markedly lower than the costs associated with other methods, which are 3.7 to 7.5 times higher. This superior performance in both accuracy and cost-effectiveness validates our framework's ability to provide reliable, efficient, and economically viable fault detection for lithium-ion batteries.

\section{Conclusion}
In this work, we introduced \textbf{BatteryBERT}, a novel framework that successfully adapts the pre-training paradigm of LLMs for the challenging task of realistic battery fault detection. By leveraging a specialized time-series embedding module and a point-MSM self-supervised pre-training task, our approach effectively captures the complex temporal dynamics inherent in battery operational data. We demonstrated that initializing our model with parameters pre-trained on natural language provides a significant advantage, accelerating convergence and improving the stability of the pre-training process.

Our experimental results validate the efficacy of this approach. The t-SNE visualizations clearly show that \textbf{BatteryBERT} learns to generate distributionally robust, vehicle-agnostic representations, effectively normalizing the data distributions from different vehicles into a common feature space. This superior representational quality translates directly into state-of-the-art performance on the downstream fault detection task. Our framework achieved an AUROC of 94.5\%, substantially outperforming existing methods. Furthermore, the practical benefits were highlighted by a substantial reduction in the average direct cost of faults to just 229 CNY, showcasing the model's high precision and economic viability.

In conclusion, our research confirms that BERT’s architectural principles and self-supervised learning strategies can be repurposed for industrial time-series analysis. The \textbf{BatteryBERT} framework provides a scalable, accurate, and cost-effective solution for enhancing the safety and reliability of lithium-ion batteries in electric vehicles and paves the way for broader applications of LLM-inspired architectures in predictive maintenance and other critical industrial domains.



\begin{thebibliography}{00}
\bibitem{ref6}Zhang J, Wang Y, Jiang B, et al. Realistic fault detection of li-ion battery via dynamical deep learning[J]. Nature Communications, 2023, 14(1): 5940.
\bibitem{ref7}Sun G, Wang X, Zhang X, et al. Autoencoder-enhanced regularized prototypical network for new energy vehicle battery fault detection[J]. Control Engineering Practice, 2023, 141: 105738.
\bibitem{ref8}Lai X, Yu J, Mao S, et al. Enhancing early warning systems: Experimental investigation of physical signals in thermal runaway evolution of large-capacity lithium iron phosphate batteries[J]. Journal of Power Sources, 2025, 632: 236389.
\bibitem{ref9}Cao R, Zhang Z, Shi R, et al. Model-constrained deep learning for online fault diagnosis in Li-ion batteries over stochastic conditions[J]. Nature Communications, 2025, 16(1): 1651.
\bibitem{ref10}He H, Zhao X, Li J, et al. Voltage abnormality-based fault diagnosis for batteries in electric buses with a self-adapting update model[J]. Journal of Energy Storage, 2022, 53: 105074.
\bibitem{ref11}Zhang C, Li S, Du J, et al. Graph-guided fault detection for multi-type lithium-ion batteries in realistic electric vehicles optimized by ensemble learning[J]. Journal of Energy Chemistry, 2025, 106: 507-522.

\bibitem{ref5}Zhang Z, Zhu Y, Zhang Q, et al. Multi-cycle charging information guided state of health estimation for lithium-ion batteries based on pre-trained large language model[J]. Energy, 2024, 313: 133993.

\bibitem{min2023recent}Min B, Ross H, Sulem E, et al. Recent advances in natural language processing via large pre-trained language models: A survey[J]. ACM Computing Surveys, 2023, 56(2): 1-40.

\bibitem{ref1}Zhao S, Chen S, Zhou J, et al. Potential to transform words to watts with large language models in battery research[J]. Cell Reports Physical Science, 2024, 5(3).
\bibitem{ref2}Peng H, Liu C, Li H. Large Language Model Enabled Health Management for Internet of Batteries in Electric Vehicles[J]. IEEE Internet of Things Journal, 2024.
\bibitem{ref3}Bian C, Han X, Duan Z, et al. Hybrid prompt-driven large language model for robust state-of-charge estimation of multi-type li-ion batteries[J]. IEEE Transactions on Transportation Electrification, 2024.
\bibitem{ref4}Bian C, Duan Z, Hao Y, et al. Exploring large language model for generic and robust state-of-charge estimation of Li-ion batteries: A mixed prompt learning method[J]. Energy, 2024, 302: 131856.

\bibitem{ref_bert_variant}Kenton J D M W C, Toutanova L K. Bert: Pre-training of deep bidirectional transformers for language understanding[C]. Proceedings of NAACL-HLT, 2019, 1: 2.

\bibitem{ke2017lightgbm} Ke G, Meng Q, Finley T, et al. Lightgbm: A highly efficient gradient boosting decision tree[J]. Advances in Neural Information Processing Systems, 2017, 30.


\bibitem{cheng2021supervised} Cheng Y, Wang X, Xia Y. Supervised t-distributed stochastic neighbor embedding for data visualization and classification[J]. INFORMS Journal on Computing, 2021, 33(2).

\bibitem{deng2021graph} Deng A, Hooi B. Graph neural network-based anomaly detection in multivariate time series[C]. Proceedings of the AAAI Conference on Artificial Intelligence, 2021, 35(5), pp. 4027--4035.

\bibitem{stent2005evaluating} Stent A, Marge M, Singhai M. Evaluating evaluation methods for generation in the presence of variation[C]. International Conference on Intelligent Text Processing and Computational Linguistics, 2005, pp. 341--351.



\end{thebibliography}
\end{document}